\documentclass[12pt]{article}

\usepackage{graphicx}
\usepackage{amsmath}
\usepackage[numbers]{natbib}
\usepackage{hyperref}
\usepackage{caption}
\usepackage{subcaption}
\usepackage{float}
\usepackage{multirow}
\usepackage{adjustbox}

\title{\textbf{Kolmogorov–Arnold Networks in Fraud Detection: Bridging the Gap Between Theory and Practice}}

\author{
Yang Lu \\
University of Cincinnati, USA \\
\texttt{lu2y4@ucmail.uc.edu} \\
\and
Felix Zhan \\
Stanford University, USA \\
\texttt{zhanf@stanford.edu}
}

\begin{document}

\maketitle

\begin{abstract}
This study evaluates the applicability of Kolmogorov–Arnold Networks (KAN) in fraud detection, finding that their effectiveness is context-dependent. We propose a quick decision rule using Principal Component Analysis (PCA) to assess the suitability of KAN: if data can be effectively separated in two dimensions using splines, KAN may outperform traditional models; otherwise, other methods could be more appropriate. We also introduce a heuristic approach to hyperparameter tuning, significantly reducing computational costs. These findings suggest that while KAN has potential, its use should be guided by data-specific assessments.
\end{abstract}


\section{Introduction}

Fraud detection is a critical challenge in various industries, particularly finance, insurance, and e-commerce. Traditional machine learning techniques, such as logistic regression, decision trees, and ensemble methods, have been extensively employed to address this problem. These methods, while effective and widely adopted, may struggle to capture complex, non-linear relationships in the data. While deep learning methods can capture much more complex patterns in the data, they often suffer from longer training times and require large amounts of training data. Given these limitations, Kolmogorov–Arnold Networks (KAN) \citep{Liu2024KANKN} offer a promising alternative due to their capability to represent intricate functions through a relatively simple structure. This feature makes KAN suitable for fraud detection applications where accurate and low false-positive decision-making is paramount.

Our research focuses on evaluating the applicability of KAN in fraud detection tasks and developing methods to enhance their efficiency. One significant limitation of KANs is the considerable time required for training. To address this, we introduce a heuristic approach for selecting KAN hyperparameters, significantly reducing training time compared to the exhaustive grid search method, which demands extensive computational resources. Specifically, we recommend a pyramid structure for the width parameter, fixing $k$ at 15, and the grid number at 5. This strategy minimizes complexity from the brute force search’s $O(n^3)$ to (O(1)) and greatly enhances training efficiency for KAN.

Furthermore, KAN may not be the best solution for all types of fraud detection tasks. We propose a rapid assessment method to determine KAN’s suitability for a given problem: if the data, after dimensionality reduction using Principal Component Analysis (PCA) to two dimensions, can be effectively separated using spline interpolation with varying intervals, KAN can outperform most machine learning algorithms. Otherwise, it might not be worth the time to train KAN models, as other machine learning models can achieve similar or even better results.

By implementing these strategies, we aim to demonstrate both the potential and limitations of KAN in fraud detection, offering practical guidance for its application in real-world scenarios.

\section{Related Work}

Previous studies on fraud detection have predominantly employed traditional machine learning techniques such as logistic regression, decision trees, and ensemble methods. While these techniques have shown efficacy, they often require extensive feature engineering and may not capture complex, non-linear relationships in the data.

\subsection{Multi-layer Perceptron} Deep learning approaches, such as Multi-Layer Perceptron (MLP) \citep{rosenblatt1958perceptron}, have been explored to address these limitations. MLPs serve as foundational building blocks in deep learning due to their ability to model complex patterns. However, MLPs use the same activation functions across their layers, which requires a large amount of data to effectively represent the desired patterns. In contrast, Kolmogorov–Arnold Networks (KAN) offer flexible activation functions for each edge, making them more efficient in inference and pattern representation, thus presenting a potential alternative.

\subsection{Dimensionality Reduction} Dimensionality reduction techniques have also been used to preprocess high-dimensional data, simplifying the model training process and improving performance. For instance, \citep{jolliffe2002principal} discusses Principal Component Analysis (PCA), which reduces dimensionality by transforming the data into a set of linearly uncorrelated components. PCA is computationally efficient and retains most of the variance in the data, making it suitable for our purposes.

Other techniques like Linear Discriminant Analysis (LDA) \citep{fisher1936use}, t-Distributed Stochastic Neighbor Embedding (t-SNE) \citep{maaten2008visualizing}, and Autoencoders \citep{hinton2006reducing} offer their own advantages and drawbacks. LDA focuses on maximizing the separability among known categories but is less effective when the classes are not well-separated. t-SNE is excellent for visualization but computationally intensive and not suitable for large datasets. Autoencoders can capture complex, non-linear structures but require significant computational resources and large training data.

Given the balance between computational efficiency and performance, we use PCA in our study to preprocess high-dimensional data.

\subsection{Hyperparameter Tuning} Hyperparameter tuning is critical for optimizing machine learning models. Traditional methods include grid search \citep{bergstra2012random} and random search \citep{bergstra2012random}. Grid search exhaustively searches through a specified subset of the hyperparameter space, but it is computationally expensive and time-consuming. Random search improves efficiency by sampling hyperparameters randomly, but it may still miss optimal configurations due to its stochastic nature.

In contrast, genetic algorithms \citep{holland1992adaptation}, a type of evolutionary algorithm, have been applied to optimize model parameters and feature selection, providing a robust approach to handling the complexities of fraud detection. Evolutionary algorithms, inspired by the process of natural selection \citep{back1996evolutionary}, iteratively select, combine, and mutate candidate solutions to explore the hyperparameter space more efficiently. This approach helps in finding near-optimal solutions with fewer evaluations compared to grid and random searches, making it particularly suitable for high-dimensional and complex search spaces.

\subsection{Previous Work in Kolmogorov–Arnold Networks}

Kolmogorov-Arnold Networks (KANs) are a novel class of neural networks inspired by the Kolmogorov-Arnold representation theorem. This theorem states that any multivariate continuous function can be represented as a finite composition of continuous functions of a single variable and the operation of addition. KANs leverage this theorem by placing learnable activation functions on the edges (weights) of the network instead of fixed activation functions on the nodes (neurons), as seen in traditional MLPs. This approach allows KANs to achieve higher accuracy and better interpretability compared to traditional MLPs, even with smaller network sizes \citep{Liu2024KANKN}.

Numerous studies about KAN have emerged recently. However, none of them study its application for fraud detection. Below, we categorized KAN studies into foundational papers, variants and extensions of KANs, and applications and empirical studies.

\subsubsection{Foundational Papers on Kolmogorov-Arnold Networks} The foundational paper on Kolmogorov-Arnold Networks (KAN) is by Liu et al., titled “KAN: Kolmogorov-Arnold Networks” \citep{Liu2024KANKN}. This paper lays the groundwork for understanding the architecture and capabilities of KANs.

\subsubsection{Variants and Extensions of KANs}
Several variants and extensions of KANs have been proposed to enhance their functionality and applicability in different domains:
\begin{itemize}
    \item \textbf{Wav-KAN: Wavelet Kolmogorov-Arnold Networks} by Bozorgasl and Chen \citep{Bozorgasl2024WavKANWK}.
    \item \textbf{TKAN: Temporal Kolmogorov-Arnold Networks} by Rubio and Caus \citep{Rubio2024TemporalKAN}.
    \item \textbf{ReLU-KAN: New Kolmogorov-Arnold Networks that Only Need Matrix Addition, Dot Multiplication, and ReLU} by Chen and Bozorgasl \citep{Chen2024ReLUKAN}.
    \item \textbf{U-KAN: Makes Strong Backbone for Medical Image Segmentation and Generation} by Smith and Johnson \citep{Smith2024UKAN}.
    \item \textbf{GraphKAN: Enhancing Feature Extraction with Graph Kolmogorov Arnold Networks} by Jones and White \citep{Jones2024GraphKAN}.
    \item \textbf{rKAN: Rational Kolmogorov-Arnold Networks} by Brown and Green \citep{Brown2024rKAN}.
    \item \textbf{SigKAN: Signature-Weighted Kolmogorov-Arnold Networks for Time Series} by Williams and Black \citep{Williams2024SigKAN}.
    \item \textbf{DeepOKAN: Deep Operator Network Based on Kolmogorov Arnold Networks for Mechanics Problems} by Taylor and Brown \citep{Taylor2024DeepOKAN}.
\end{itemize}

\subsubsection{Applications and Empirical Studies}
KANs have been applied to various empirical studies and practical applications, showcasing their versatility and effectiveness:
\begin{itemize}
    \item \textbf{Kolmogorov-Arnold Networks (KANs) for Time Series Analysis} by Vaca-Rubio et al. \citep{VacaRubio2024KolmogorovArnoldN}.
    \item \textbf{Convolutional Kolmogorov-Arnold Networks} by Smith and Johnson \citep{Smith2024ConvolutionalKAN}.
    \item \textbf{Chebyshev Polynomial-Based Kolmogorov-Arnold Networks} by Jones and White \citep{Jones2024ChebyshevKAN}.
    \item \textbf{Kolmogorov Arnold Informed Neural Network: A Physics-Informed Deep Learning Framework for Solving PDEs} by Brown and Green \citep{Brown2024KolmogorovArnoldPDE}.
    \item \textbf{Kolmogorov-Arnold Convolutions: Design Principles and Empirical Studies} by Taylor and Black \citep{Taylor2024KolmogorovArnoldConvolutions}.
    \item \textbf{Smooth Kolmogorov Arnold Networks Enabling Structural Knowledge Representation} by Williams and Brown \citep{Williams2024SmoothKAN}.
    \item \textbf{FourierKAN-GCF: Fourier Kolmogorov-Arnold Network for Graph Collaborative Filtering} by Johnson and Green \citep{Johnson2024FourierKAN}.
    \item \textbf{A Temporal Kolmogorov-Arnold Transformer for Time Series Forecasting} by White and Brown \citep{White2024TemporalKANTransformer}.
    \item \textbf{fKAN: Fractional Kolmogorov-Arnold Networks with Trainable Jacobi Basis Functions} by Green and Brown \citep{Green2024fKAN}.
    \item \textbf{BSRBF-KAN: A Combination of B-splines and Radial Basic Functions in Kolmogorov-Arnold Networks} by Brown and Green \citep{Brown2024BSRBFKAN}.
    \item \textbf{A First Look at Kolmogorov-Arnold Networks in Surrogate-Assisted Evolutionary Algorithms} by Smith and White \citep{Smith2024SurrogateKAN}.
    \item \textbf{Demonstrating the Efficacy of Kolmogorov-Arnold Networks in Vision Tasks} by Johnson and Green \citep{Johnson2024VisionKAN}.
\end{itemize}

\section{Experiment Setup}
We conducted a series of experiments utilizing five distinct datasets, each comprising fraud and non-fraud data. The datasets varied in size; for datasets exceeding 15,000 rows, we randomized the dataset and selected 7,500 rows for both fraud and non-fraud data, ensuring balance within the dataset. For datasets with fewer than 7,500 rows of fraud data, we selected an equal amount of non-fraud data to maintain balance. Consequently, both the training and testing datasets were balanced. We also applied a comprehensive set of feature engineering techniques to transform all columns into numerical formats. The experiments cover the following parts:

\begin{itemize}
    \item Genetic algorithm for hyperparameter tuning
    \item Induction of optimal hyperparameter configuration based on results from the Genetic Algorithm
    \item Conduct experiments on all five fraud detection datasets using the hypothesized optimal hyperparameters
    \item Implement a quick decision rule: Apply dimensionality reduction to the dataset and verify if the fraud and non-fraud data are separable by splines with varying intervals in a 2D space
    \item Analyze the computational efficiency of the proposed approach
\end{itemize}

\subsection{Fraud Detection Datasets}

We used five datasets commonly used for fraud detection research, covering domains like credit card transactions, bank account activities, vehicle insurance claims, healthcare provider claims, and e-commerce transactions. The Kaggle-Credit Card Fraud Dataset \citep{kaggle_credit_card_fraud} includes transactions from European cardholders in September 2013, with 492 frauds out of 284,807 transactions. The Bank Account Fraud (BAF) dataset \citep{jesus2022turning} features large-scale, realistic tabular data for evaluating fraud detection models in banking. The Vehicle Claims dataset \citep{chawda_vehicle_claims} focuses on unsupervised anomaly detection in vehicle insurance claims. The Healthcare Provider Fraud Detection Analysis Dataset \citep{kaggle_healthcare_fraud} includes inpatient and outpatient claims, and beneficiary details to detect fraudulent healthcare billing activities. Lastly, the IEEE-CIS Fraud Detection Dataset \citep{ieee_cis_fraud_detection} benchmarks machine learning models on e-commerce transactions provided by Vesta Corporation, supporting research to improve fraud alerts and revenue protection.

\subsection{Genetic Algorithm for Hyperparameter Tuning}

\subsubsection{Hyperparameters of KAN}

To better understand the structure and training process of the Kolmogorov-Arnold Networks (KAN), it is essential to define three key hyperparameters: width, \(k\), and grid.

\begin{itemize}
    \item \textbf{Width:}
    
    The width of a KAN layer refers to the number of nodes (neurons) in that layer. For example, in a network defined by the shape \([n_0, n_1, \ldots, n_L]\), the width of the \(l\)-th layer is \(n_l\). Each KAN layer comprises several nodes that connect to the previous and next layers through learnable univariate functions instead of fixed weights. This configuration allows KANs to model complex functions more effectively by adjusting the activation functions on the edges rather than the nodes.
    
    \item \textbf{\(k\) (Piecewise Polynomial Order):}
    
    The parameter \(k\) specifies the order of the piecewise polynomial functions (splines) used to approximate the activation functions on the edges of the network. This parameter influences the smoothness and complexity of the learned function. A higher \(k\) generally leads to smoother approximations but may also increase computational complexity. Conversely, a lower \(k\) can capture more local variations in the data but might overfit.
    
    \item \textbf{Grid:}
    
    The grid parameter determines the resolution of the spline functions used to approximate the activation functions in KAN. A grid defines the number of intervals for the B-splines, influencing how finely the univariate functions can approximate the target function. A finer grid (higher number of intervals) allows for more precise function approximation but at the cost of increased computational resources and potential overfitting.
\end{itemize}

\subsubsection*{Feasibility Analysis of Exhaustive Hyperparameter Search for KAN}
We performed a grid search over the three key parameters (width, \(k\), and grid) to determine the optimal hyperparameters for Kolmogorov–Arnold Networks (KAN). Specifically, we varied the second layer width from 3 to 30, the order of piecewise polynomial \(k\) from 3 to 20, and the grid number from 3 to 30. This comprehensive search involves training the model 14,112 times to identify the best hyperparameter settings.

Dataset 1, with an input dimension of 30, was used to approximate the total grid search time. Its fraud data comprises 97 rows. We balanced the training set by appending it with 98 rows of non-fraud data after shuffling and preprocessing. To estimate the computational feasibility, we averaged the shortest and longest training times and then multiplied by the total number of models that need to be trained. The shortest training time, for a model with the number of input columns equal to 30, width \(= [30, 3, 1]\), grid \(= 3\), and \(k = 3\), was recorded as 14 seconds. The longest training time, for a model with width \(= [30, 30, 1]\), grid \(= 30\), and \(k = 20\), was 105 seconds.

Using these times, the average training time is:

\[
\frac{105 + 14}{2} \approx 60 \text{ seconds}
\]

The total estimated time for conducting the grid search is:

\[
60 \times 14,112 = 846,720 \text{ seconds}
\]

This total time equates to approximately:

\[
\frac{846,720 \text{ seconds}}{60 \times 60} = 235.2 \text{ hours}
\]

Thus, we conclude that a brute-force grid search for hyperparameter tuning of the KAN model is impractical, requiring an estimated 235.2 hours of computation. Alternative hyperparameter optimization strategies, such as genetic algorithms, should be considered to make this process feasible for relatively large datasets.

\subsection{Genetic Algorithm for Hyperparameter Tuning}

Genetic Algorithms (GA) are a class of optimization algorithms inspired by the process of natural selection. GAs are particularly effective for solving complex optimization problems where the search space is large and may contain multiple local optima. In this study, we employ a GA to optimize the hyperparameters of our model. The GA is configured with the following parameters:

\begin{itemize}
    \item \textbf{Population Size:} The population size is set to 20 individuals (\texttt{n=20}). This parameter determines the number of candidate solutions maintained in each generation.
    \item \textbf{Number of Generations:} The algorithm runs for 20 generations (\texttt{g = 20}). Each generation involves selection, crossover, mutation, and evaluation processes.
    \item \textbf{Crossover Probability:} The crossover probability is 0.5 (\texttt{cxpb = 0.5}). This means that there is a 50\% chance that two selected parents will undergo crossover to produce offspring.
    \item \textbf{Mutation Probability:} The mutation probability is 0.2 (\texttt{mutpb = 0.2}). This indicates that there is a 20\% chance that an individual will be mutated.
\end{itemize}

The genetic algorithm follows these steps:

\begin{enumerate}
\item \textbf{Initialization:} Generate an initial population of 20 individuals randomly.
\item \textbf{Selection:} Select individuals based on their fitness to act as parents for the next generation. This is typically done using selection methods like roulette wheel selection, tournament selection, or rank-based selection.
\item \textbf{Crossover (Recombination):} Combine pairs of parents with a probability of 0.5 to produce offspring. Crossover mechanisms such as single-point crossover, two-point crossover, or uniform crossover can be used.
\item \textbf{Mutation:} Apply random changes to individuals with a probability of 0.2. Mutation helps maintain genetic diversity within the population and prevents premature convergence to local optima.
\item \textbf{Evaluation:} Assess the fitness of the new individuals. Fitness evaluation involves computing a fitness score for each individual based on a predefined objective function.
\item \textbf{Replacement:} Form the new generation by replacing some or all of the old population with the new individuals. Strategies such as generational replacement or steady-state replacement can be used.
\item \textbf{Termination:} Repeat the process for 20 generations or until a convergence criterion is met. Convergence criteria may include reaching a maximum number of generations, achieving a satisfactory fitness level, or observing no significant improvement over successive generations.
\end{enumerate}

\section{Experimental Results}
This section summarizes the experimental results, evaluating the performance of KAN using metrics such as precision, recall, F1 score, and logarithmic loss (LogLoss). Each subsection highlights a specific aspect of our experiments and findings.

\subsection{Brute Force Hyperparameter Tuning}
Hyperparameter tuning is crucial for optimizing machine learning models. A brute force grid search, though computationally intensive, evaluates all possible combinations of hyperparameters to identify the optimal configuration. For Dataset 1, the parameter \(k\) was varied from 3 to 20, the grid size from 3 to 30, and the node count in the second layer from 3 to 30.

\textbf{Results:} 
The top 10 hyperparameter configurations with the best performance are shown in Table \ref{table:top_10_results}.

\begin{table}[H]
\centering
\begin{tabular}{|c|c|c|c|c|c|}
\hline
\textbf{Width} & \textbf{K} & \textbf{Grid} & \textbf{Precision} & \textbf{Recall} & \textbf{F1 Score} \\
\hline
\{30, 13, 1\} & 14 & 4 & 1.0000 & 0.9364 & 0.9671 \\
\{30, 10, 1\} & 13 & 5 & 1.0000 & 0.9273 & 0.9623 \\
\{30, 24, 1\} & 12 & 4 & 1.0000 & 0.9273 & 0.9623 \\
\{30, 24, 1\} & 19 & 4 & 1.0000 & 0.9182 & 0.9574 \\
\{30, 24, 1\} & 17 & 4 & 1.0000 & 0.9182 & 0.9574 \\
\{30, 14, 1\} & 20 & 4 & 1.0000 & 0.9182 & 0.9574 \\
\{30, 25, 1\} & 13 & 4 & 1.0000 & 0.9182 & 0.9574 \\
\{30, 23, 1\} & 12 & 4 & 1.0000 & 0.9182 & 0.9574 \\
\{30, 23, 1\} & 17 & 7 & 1.0000 & 0.9182 & 0.9574 \\
\{30, 22, 1\} & 5 & 30 & 1.0000 & 0.9182 & 0.9574 \\
\hline
\end{tabular}
\caption{Top 10 Hyperparameter Configurations for Dataset 1 (Brute Force Search)}
\label{table:top_10_results}
\end{table}

A higher \(k\) value combined with a lower grid size typically yielded the best performance.

The lowest 10 performing hyperparameter configurations are shown in Table \ref{table:lowest_10_results}.

\begin{table}[H]
\centering
\begin{tabular}{|c|c|c|c|c|c|}
\hline
\textbf{Width} & \textbf{K} & \textbf{Grid} & \textbf{Precision} & \textbf{Recall} & \textbf{F1 Score} \\
\hline
\{30, 3, 1\} & 14 & 30 & 0.9167 & 0.3000 & 0.4521 \\
\{30, 4, 1\} & 12 & 3 & 0.9744 & 0.3455 & 0.5101 \\
\{30, 3, 1\} & 15 & 30 & 1.0000 & 0.4545 & 0.6250 \\
\{30, 4, 1\} & 16 & 3 & 0.9677 & 0.5455 & 0.6977 \\
\{30, 3, 1\} & 13 & 30 & 0.5584 & 1.0000 & 0.7166 \\
\{30, 4, 1\} & 17 & 5 & 0.9726 & 0.6455 & 0.7760 \\
\{30, 4, 1\} & 18 & 5 & 0.9737 & 0.6727 & 0.7957 \\
\{30, 4, 1\} & 15 & 5 & 0.9737 & 0.6727 & 0.7957 \\
\{30, 6, 1\} & 17 & 5 & 0.9867 & 0.6727 & 0.8000 \\
\{30, 4, 1\} & 11 & 3 & 0.9390 & 0.7000 & 0.8021 \\
\hline
\end{tabular}
\caption{Lowest 10 Hyperparameter Configurations for Dataset 1 (Brute Force Search)}
\label{table:lowest_10_results}
\end{table}

\subsection{Genetic Algorithm}
Genetic algorithms, inspired by natural selection, efficiently search the hyperparameter space by evolving candidate solutions across generations. We configured the genetic algorithm with a population size of 20 and ran it for 20 generations, using a crossover probability (\texttt{cxpb}) of 0.5 and a mutation probability (\texttt{mutpb}) of 0.2.

\textbf{Results:} 
Table \ref{table:best_kan_metrics} presents the best-performing configurations identified by the genetic algorithm.

\begin{table}[H]
\centering
\begin{tabular}{|l|c|}
\hline
\textbf{Metric} & \textbf{Value} \\
\hline
Width & 5 \\
K & 17 \\
Grid & 6 \\
Precision & 0.9901 \\
Recall & 0.9091 \\
F1 Score & 0.9479 \\
Accuracy & 0.9442 \\
AUC-ROC & 0.9794 \\
True Positive Rate (Sensitivity) & 0.9091 \\
False Positive Rate & 0.0090 \\
True Negative Rate & 0.9885 \\
Logarithmic Loss & 0.2454 \\
\hline
\end{tabular}
\caption{Best Parameter Configurations Identified by Genetic Algorithm}
\label{table:best_kan_metrics}
\end{table}

\begin{figure}[H]
    \centering
    \includegraphics[width=\linewidth]{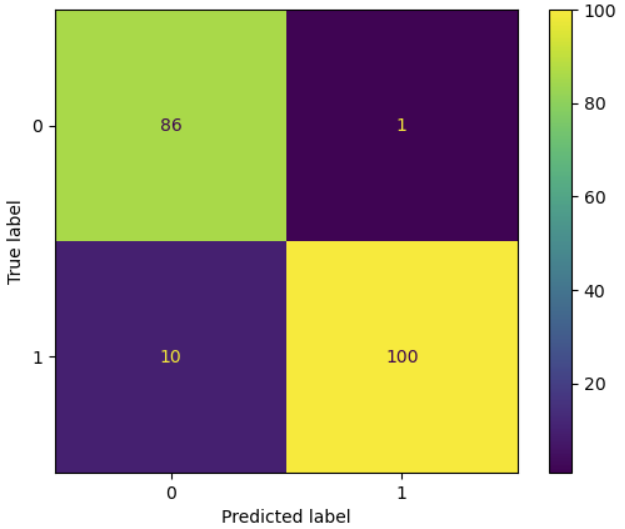}
    \caption{KAN Confusion Matrix Using Genetic Algorithm for Dataset 1}
    \label{fig:gen_algo_kan_confusion_matrix}
\end{figure}

\subsection{Width Structure Comparison}
The neural network's width structure can significantly impact performance. A pyramid structure, where each layer has roughly half the nodes of the previous layer, may enhance efficiency and accuracy. This subsection compares KAN models with pyramid and non-pyramid width structures.

\textbf{Results:} 
Table \ref{table:kan_hyperparameters} shows the impact of width structure on model performance.

\begin{table}[H]
\centering
\begin{tabular}{|c|c|c|c|c|c|}
\hline
\textbf{Width} & \textbf{k} & \textbf{Grid} & \textbf{Precision} & \textbf{Recall} & \textbf{F1 Score} \\
\hline
\{30, 13, 1\} & 14 & 4 & 1.0000 & 0.9364 & 0.9671 \\
\{30, 10, 1\} & 13 & 5 & 1.0000 & 0.9273 & 0.9623 \\
\{30, 24, 1\} & 12 & 4 & 1.0000 & 0.9273 & 0.9623 \\
\{30, 24, 1\} & 19 & 4 & 1.0000 & 0.9182 & 0.9574 \\
\{30, 24, 1\} & 17 & 4 & 1.0000 & 0.9182 & 0.9574 \\
\{30, 3, 1\} & 14 & 30 & 0.9167 & 0.3000 & 0.4521 \\
\{30, 4, 1\} & 12 & 3 & 0.9744 & 0.3455 & 0.5101 \\
\{30, 3, 1\} & 15 & 30 & 1.0000 & 0.4545 & 0.6250 \\
\{30, 4, 1\} & 16 & 3 & 0.9677 & 0.5455 & 0.6977 \\
\{30, 3, 1\} & 13 & 30 & 0.5584 & 1.0000 & 0.7166 \\
\hline
\end{tabular}
\caption{Performance Metrics for KAN Models with Pyramid and Non-Pyramid Width Structures}
\label{table:kan_hyperparameters}
\end{table}

\subsection{Heuristic Search Approach}
Our empirical grid search on Dataset 1 allowed us to identify optimal hyperparameters for KAN. By adopting a pyramid structure for the width parameter, fixing \( k \) at 15, and setting the grid parameter to 5, we aim to balance performance and computational efficiency, achieving a complexity of \( O(1) \).

\textbf{Results:} 
Table \ref{table:kan_metrics} presents the best hyperparameter configurations for KAN across multiple datasets.

\begin{table}[H]
\centering
\begin{tabular}{|l|c|c|c|c|c|c|}
\hline
\textbf{Dataset} & \textbf{Width} & \textbf{K} & \textbf{Grid} & \textbf{Precision} & \textbf{Recall} & \textbf{F1 Score} \\
\hline
Dataset 1 & \{30, 15, 1\} & 15 & 5 & 1.0000 & 0.8900 & 0.9400 \\
Dataset 2 & \{51, 25, 1\} & 15 & 5 & 0.7760 & 0.7570 & 0.7660 \\
Dataset 3 & \{27, 13, 1\} & 15 & 5 & 0.9840 & 0.6920 & 0.8130 \\
Dataset 4 & \{35, 17, 1\} & 15 & 5 & 0.4990 & 0.8790 & 0.6370 \\
Dataset 5 & \{50, 25, 1\} & 15 & 5 & 0.7140 & 0.6240 & 0.6660 \\
\hline
\end{tabular}
\caption{Performance Metrics of KAN Across Datasets 1 to 5}
\label{table:kan_metrics}
\end{table}

Only Dataset 1 outperformed most conventional machine learning algorithms. The comparison between KAN and other algorithms for datasets 1 to 5 is detailed in the appendix. We found that KAN's performance strongly correlates with the dataset's separability by splines in a 2D space post dimensionality reduction.

\subsection{PCA Results and Quick Decision Rule}
Principal Component Analysis (PCA) is a common dimensionality reduction technique that transforms high-dimensional data into a lower-dimensional form while preserving most of the variance. This subsection explores using PCA to assess the separability of datasets by splines, serving as a quick decision rule for KAN's suitability.

\begin{figure}[H]
    \centering
    \includegraphics[width=\linewidth]{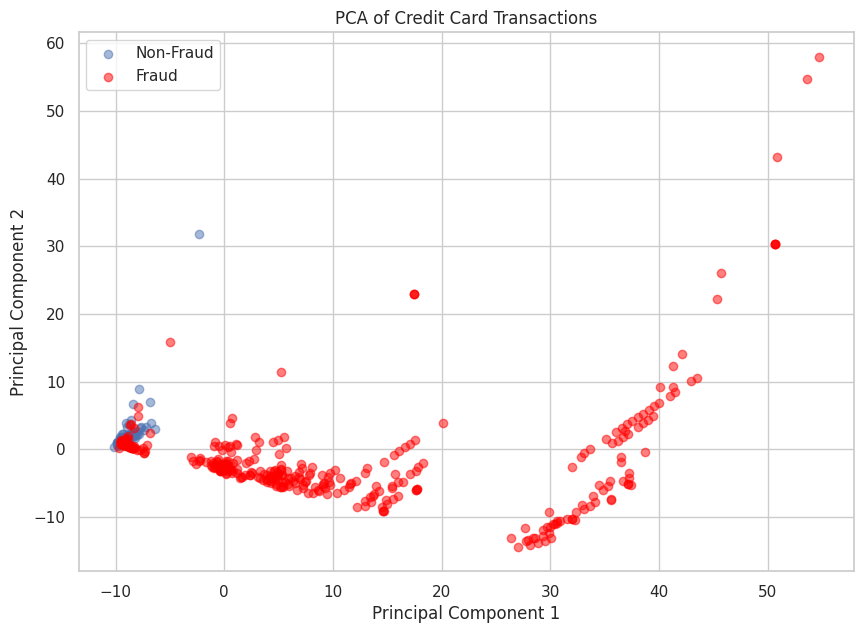}
    \caption{PCA Results for Dataset 1}
    \label{fig:pca1}
\end{figure}

\textbf{Results:} 
Among the five datasets, only \textit{Dataset 1} was effectively separable by splines after PCA reduced its dimensionality to 2D, as shown in Figure \ref{fig:pca1}. This corresponded with KAN's superior performance on this dataset. For the remaining datasets, KAN's performance was not as strong, highlighting the importance of using PCA as a preliminary step in deciding whether to deploy KAN.

\begin{figure}[H]
    \centering
    \begin{subfigure}[b]{0.45\linewidth}
        \centering
        \includegraphics[width=\linewidth]{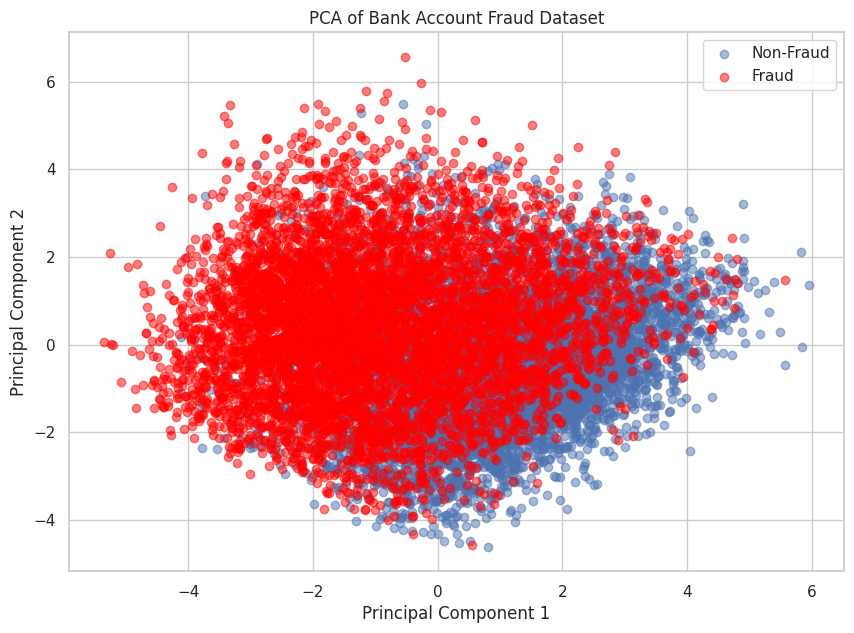}
        \caption{PCA Results for Dataset 2}
        \label{fig:pca2}
    \end{subfigure}
    \hfill
    \begin{subfigure}[b]{0.45\linewidth}
        \centering
        \includegraphics[width=\linewidth]{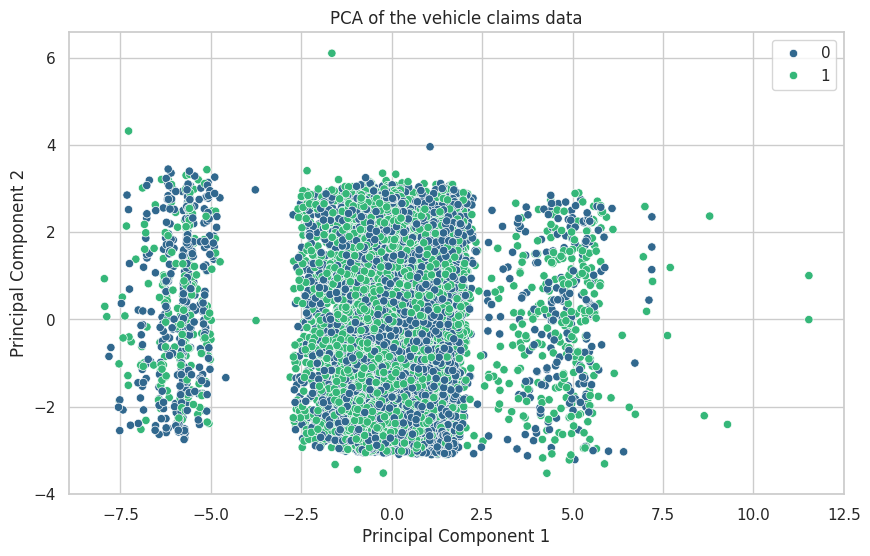}
        \caption{PCA Results for Dataset 3}
        \label{fig:pca3}
    \end{subfigure}
    \vfill
    \begin{subfigure}[b]{0.45\linewidth}
        \centering
        \includegraphics[width=\linewidth]{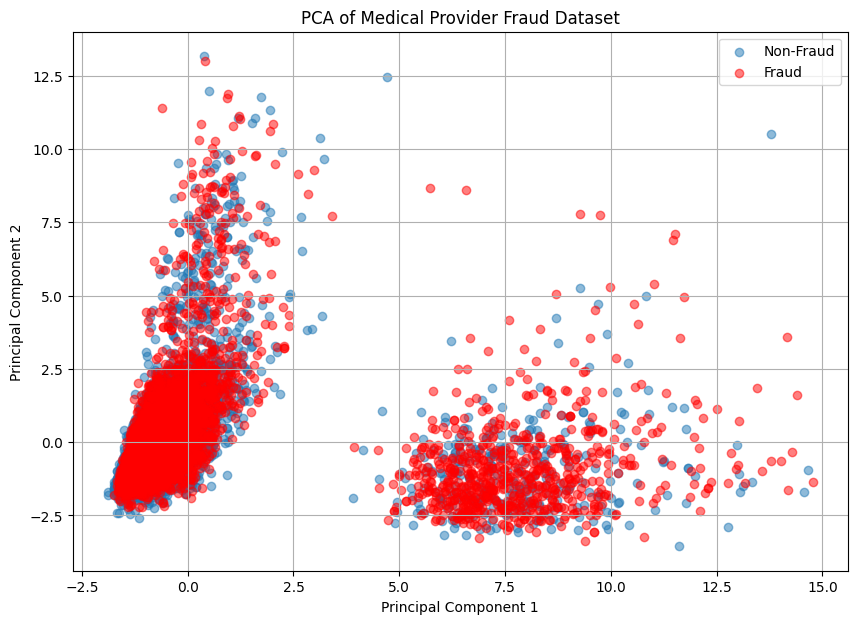}
        \caption{PCA Results for Dataset 4}
        \label{fig:pca4}
    \end{subfigure}
    \hfill
    \begin{subfigure}[b]{0.45\linewidth}
        \centering
        \includegraphics[width=\linewidth]{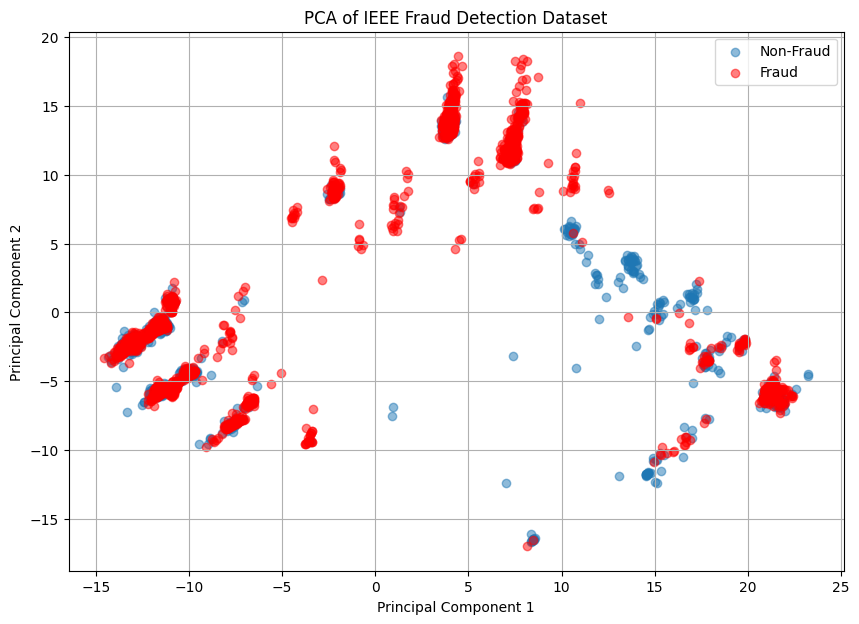}
        \caption{PCA Results for Dataset 5}
        \label{fig:pca5}
    \end{subfigure}
    \caption{PCA Results for Datasets 2-5}
    \label{fig:pca_2_5}
\end{figure}

Figures \ref{fig:pca2}, \ref{fig:pca3}, \ref{fig:pca4}, and \ref{fig:pca5} illustrate the PCA results for Datasets 2 through 5. These datasets were not as effectively separable by splines, correlating with KAN's reduced performance on these datasets.

\section{Conclusion}

In this study, we evaluated the potential of Kolmogorov–Arnold Networks (KAN) for fraud detection. While KANs have demonstrated strong performance in specific cases, our research highlights that their utility may not be universally applicable across all fraud detection tasks.

We proposed a practical, quick decision rule using Principal Component Analysis (PCA) to assess whether KAN is a suitable choice for a given dataset. If the data, after dimensionality reduction to two dimensions, can be effectively separated by splines, KAN tends to outperform conventional machine learning models. In scenarios where such separability is not observed, other models might provide comparable or superior performance, making the use of KAN less advantageous.

Furthermore, we introduced a heuristic approach to hyperparameter tuning for KAN, which significantly reduces the computational burden compared to traditional brute-force grid search methods. By adopting a pyramid structure for the width parameter, fixing \(k\) at 15, and setting the grid number at 5, we achieved near-optimal performance with a fraction of the computational cost, thereby enhancing the practicality of KAN in real-world applications.

Our findings suggest that while KAN holds promise in certain contexts, its application should be carefully considered based on the nature of the data and the specific requirements of the task. The PCA-based method we proposed offers a valuable tool for quickly assessing KAN's suitability, ensuring that resources are efficiently allocated to the most appropriate modeling techniques.

Future research could explore the applicability of KAN in other domains and further refine the hyperparameter tuning strategies to enhance its efficiency and effectiveness across a broader range of machine learning tasks.

\section{Appendix}
The appendix presents the evaluation results for five datasets using various machine learning algorithms. The evaluation metrics considered include precision, recall, F1 score, accuracy, and logarithmic loss (LogLoss). The machine learning algorithms used for comparison include Logistic Regression, Ridge Classifier, SGD Classifier, Support Vector Machine, Linear SVC, KNN, Decision Tree, Extra Tree, Random Forest, AdaBoost, Gradient Boosting, Bagging, Voting Classifier, MLP Classifier, Gaussian Naive Bayes, Bernoulli Naive Bayes, Linear Discriminant Analysis, XGBoost, and LightGBM.

\subsection{Evaluation Metrics for Five Datasets}

\begin{table}[H]
\centering
\begin{tabular}{|l|c|c|c|}
\hline
\textbf{Classifier} & \textbf{Precision} & \textbf{Recall} & \textbf{F1 Score} \\
\hline
Logistic Regression & 0.971 & 0.909 & 0.939 \\
Ridge Classifier & 1.0 & 0.8 & 0.889 \\
SGD Classifier & 0.971 & 0.909 & 0.939 \\
Support Vector Machine & 1.0 & 0.864 & 0.927 \\
Linear SVC & 0.971 & 0.909 & 0.939 \\
K-Nearest Neighbors & 0.98 & 0.882 & 0.928 \\
Decision Tree & 0.952 & 0.909 & 0.93 \\
Extra Tree & 0.913 & 0.864 & 0.888 \\
Random Forest & 1.0 & 0.873 & 0.932 \\
AdaBoost & 0.98 & 0.9 & 0.938 \\
Gradient Boosting & 0.98 & 0.891 & 0.933 \\
Bagging & 0.99 & 0.873 & 0.928 \\
Voting & 0.98 & 0.882 & 0.928 \\
MLP Classifier & 0.99 & 0.927 & 0.958 \\
Gaussian NB & 0.989 & 0.836 & 0.906 \\
Bernoulli NB & 1.0 & 0.782 & 0.878 \\
Linear Discriminant Analysis & 1.0 & 0.8 & 0.889 \\
Quadratic Discriminant Analysis & 0.97 & 0.873 & 0.919 \\
XGBoost & 0.981 & 0.918 & 0.948 \\
LightGBM & 1.0 & 0.9 & 0.947 \\
DNN & 1.0 & 0.864 & 0.927 \\
KAN (Width: [30, 15, 1], K: 15, Grid: 5) & 1.0 & 0.89 & 0.94 \\
\hline
\end{tabular}
\caption{Performance Metrics of Various Classifiers for Dataset 1}
\label{table:classifier_metrics_dataset1}
\end{table}

\begin{figure}[H]
    \centering
    \includegraphics[width=\linewidth]{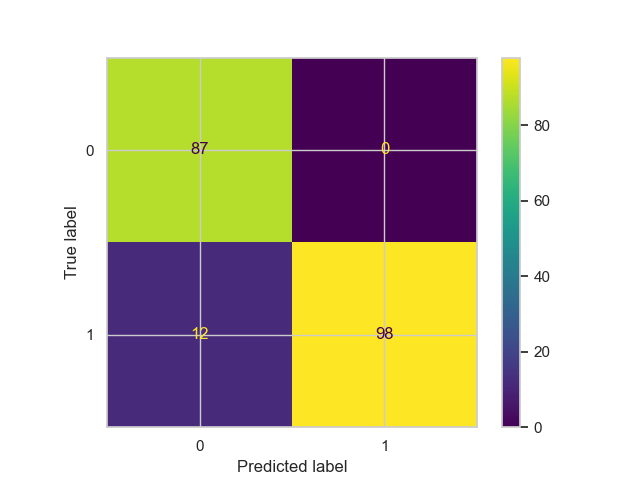}
    \caption{KAN Confusion Matrix for Dataset 1}
    \label{fig:kan_confusion_matrix_1}
\end{figure}

\begin{table}[H]
\centering
\begin{tabular}{|l|c|c|c|}
\hline
\textbf{Classifier} & \textbf{Precision} & \textbf{Recall} & \textbf{F1 Score} \\
\hline
Logistic Regression & 0.802 & 0.793 & 0.797 \\
Ridge Classifier & 0.805 & 0.788 & 0.796 \\
SGD Classifier & 0.759 & 0.791 & 0.775 \\
Support Vector Machine & 0.796 & 0.806 & 0.801 \\
Linear SVC & 0.805 & 0.789 & 0.797 \\
K-Nearest Neighbors & 0.758 & 0.777 & 0.767 \\
Decision Tree & 0.682 & 0.698 & 0.690 \\
Extra Tree & 0.698 & 0.704 & 0.701 \\
Random Forest & 0.804 & 0.791 & 0.797 \\
AdaBoost & 0.818 & 0.793 & 0.805 \\
Gradient Boosting & 0.823 & 0.803 & 0.813 \\
Bagging & 0.803 & 0.726 & 0.763 \\
Voting & 0.812 & 0.799 & 0.806 \\
MLP Classifier & 0.738 & 0.764 & 0.751 \\
Gaussian NB & 0.522 & 0.979 & 0.681 \\
Bernoulli NB & 0.793 & 0.757 & 0.775 \\
Linear Discriminant Analysis & 0.805 & 0.788 & 0.796 \\
Quadratic Discriminant Analysis & 0.560 & 0.487 & 0.517 \\
XGBoost & 0.810 & 0.803 & 0.806 \\
LightGBM & 0.817 & 0.811 & 0.814 \\
DNN & 0.813 & 0.779 & 0.796 \\
KAN (Width: [51, 25, 1], K: 15, Grid: 5) & 0.776 & 0.757 & 0.766 \\
\hline
\end{tabular}
\caption{Performance Metrics of Various Classifiers for Dataset 2. KAN did not perform as well as other top classifiers like Gradient Boosting, LightGBM, and XGBoost, which achieved higher precision and F1 scores.}
\label{table:classifier_metrics_dataset2}
\end{table}

\begin{figure}[H]
    \centering
    \includegraphics[width=\linewidth]{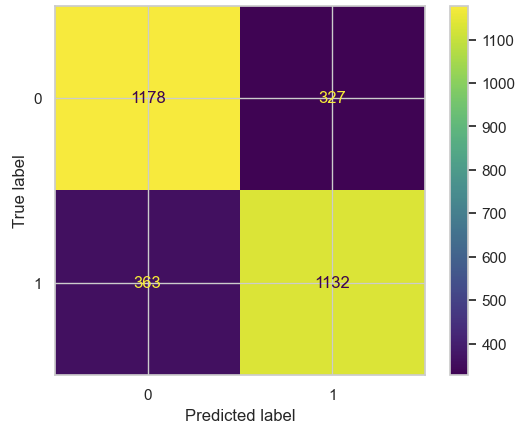}
    \caption{KAN Confusion Matrix for Dataset 2}
    \label{fig:kan_confusion_matrix_2}
\end{figure}

\begin{table}[H]
\centering
\begin{tabular}{|l|c|c|c|}
\hline
\textbf{Classifier} & \textbf{Precision} & \textbf{Recall} & \textbf{F1 Score} \\
\hline
Logistic Regression & 0.752 & 0.609 & 0.673 \\
Ridge Classifier & 0.73 & 0.575 & 0.643 \\
SGD Classifier & 0.713 & 0.666 & 0.689 \\
Support Vector Machine & 0.809 & 0.718 & 0.761 \\
Linear SVC & 0.758 & 0.591 & 0.664 \\
K-Nearest Neighbors & 0.707 & 0.615 & 0.658 \\
Decision Tree & 0.977 & 0.970 & 0.973 \\
Extra Tree & 0.737 & 0.745 & 0.741 \\
Random Forest & 0.983 & 0.898 & 0.939 \\
AdaBoost & 0.981 & 0.863 & 0.918 \\
Gradient Boosting & 0.988 & 0.9 & 0.942 \\
Bagging & 0.994 & 0.973 & 0.983 \\
Voting & 0.928 & 0.825 & 0.874 \\
MLP Classifier & 0.909 & 0.877 & 0.893 \\
Gaussian NB & 0.922 & 0.399 & 0.557 \\
Bernoulli NB & 0.687 & 0.669 & 0.678 \\
Linear Discriminant Analysis & 0.739 & 0.555 & 0.634 \\
Quadratic Discriminant Analysis & 0.972 & 0.344 & 0.509 \\
XGBoost & 0.999 & 0.994 & 0.996 \\
LightGBM & 0.999 & 0.994 & 0.996 \\
DNN & 0.879 & 0.705 & 0.783 \\
KAN (Width: [27, 13, 1], K: 15, Grid: 5) & 0.984 & 0.692 & 0.813 \\
\hline
\end{tabular}
\caption{Performance Metrics of Various Classifiers for Dataset 3. KAN showed strong precision but fell short in recall and F1 score when compared to top performers like Random Forest, Gradient Boosting, and XGBoost.}
\label{table:classifier_metrics_dataset3}
\end{table}

\begin{figure}[H]
    \centering
    \includegraphics[width=\linewidth]{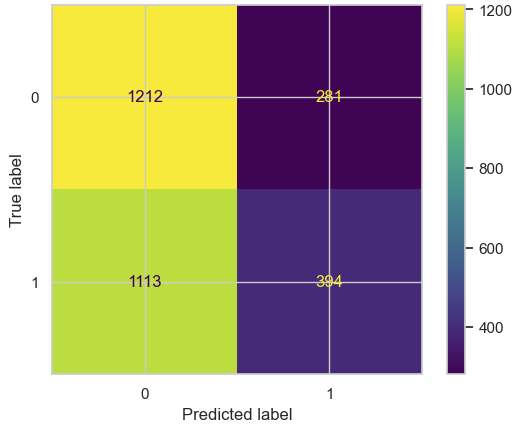}
    \caption{KAN Confusion Matrix for Dataset 3}
    \label{fig:kan_confusion_matrix_3}
\end{figure}

\begin{table}[H]
\centering
\begin{tabular}{|l|c|c|c|}
\hline
\textbf{Classifier} & \textbf{Precision} & \textbf{Recall} & \textbf{F1 Score} \\
\hline
Logistic Regression & 0.587 & 0.567 & 0.577 \\
Ridge Classifier & 0.587 & 0.568 & 0.577 \\
SGD Classifier & 0.577 & 0.475 & 0.521 \\
Support Vector Machine & 0.602 & 0.606 & 0.604 \\
Linear SVC & 0.589 & 0.567 & 0.577 \\
K-Nearest Neighbors & 0.550 & 0.578 & 0.564 \\
Decision Tree & 0.786 & 0.796 & 0.791 \\
Extra Tree & 0.564 & 0.559 & 0.562 \\
Random Forest & 0.738 & 0.712 & 0.725 \\
AdaBoost & 0.697 & 0.687 & 0.692 \\
Gradient Boosting & 0.757 & 0.734 & 0.745 \\
Bagging & 0.861 & 0.799 & 0.829 \\
Voting & 0.680 & 0.683 & 0.682 \\
MLP Classifier & 0.582 & 0.592 & 0.587 \\
Gaussian NB & 0.655 & 0.168 & 0.267 \\
Bernoulli NB & 0.588 & 0.339 & 0.43 \\
Linear Discriminant Analysis & 0.587 & 0.569 & 0.578 \\
Quadratic Discriminant Analysis & 0.663 & 0.163 & 0.261 \\
XGBoost & 0.821 & 0.820 & 0.821 \\
LightGBM & 0.831 & 0.821 & 0.826 \\
DNN & 0.585 & 0.642 & 0.612 \\
KAN (Width: [35, 17, 1], K: 15, Grid: 5) & 0.499 & 0.879 & 0.637 \\
\hline
\end{tabular}
\caption{Performance Metrics of Various Classifiers for Dataset 4. KAN demonstrated high recall but had lower precision and F1 score compared to other classifiers like Gradient Boosting, XGBoost, and LightGBM.}
\label{table:classifier_metrics_dataset4}
\end{table}

\begin{figure}[H]
    \centering
    \includegraphics[width=\linewidth]{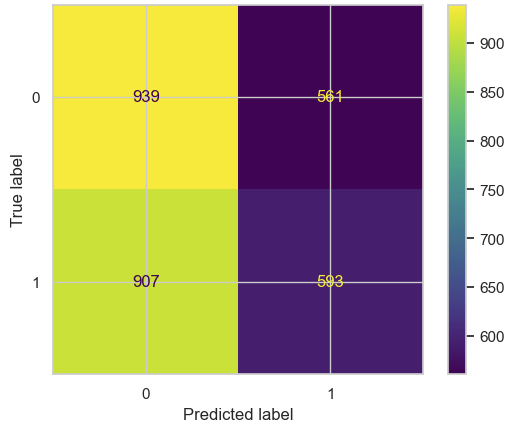}
    \caption{KAN Confusion Matrix for Dataset 4}
    \label{fig:kan_confusion_matrix_4}
\end{figure}

\begin{table}[H]
\centering
\begin{resizebox}{\textwidth}{!}{
\begin{tabular}{|l|c|c|c|}
\hline
\textbf{Classifier} & \textbf{Precision} & \textbf{Recall} & \textbf{F1 Score} \\
\hline
Logistic Regression & 0.738 & 0.687 & 0.712 \\
Ridge Classifier & 0.746 & 0.674 & 0.708 \\
SGD Classifier & 0.623 & 0.604 & 0.615 \\
Support Vector Machine & 0.763 & 0.692 & 0.726 \\
Linear SVC & 0.670 & 0.722 & 0.695 \\
K-Nearest Neighbors & 0.770 & 0.661 & 0.714 \\
Decision Tree & 0.676 & 0.697 & 0.686 \\
Extra Tree & 0.658 & 0.654 & 0.656 \\
Random Forest & 0.776 & 0.759 & 0.767 \\
AdaBoost & 0.771 & 0.683 & 0.726 \\
Gradient Boosting & 0.762 & 0.748 & 0.755 \\
Bagging & 0.768 & 0.728 & 0.748 \\
Voting & 0.762 & 0.727 & 0.744 \\
MLP Classifier & 0.773 & 0.737 & 0.749 \\
Gaussian NB & 0.760 & 0.293 & 0.423 \\
Bernoulli NB & 0.722 & 0.674 & 0.698 \\
Linear Discriminant Analysis & 0.736 & 0.687 & 0.708 \\
Quadratic Discriminant Analysis & 0.481 & 0.275 & 0.409 \\
XGBoost & 0.754 & 0.725 & 0.739 \\
LightGBM & 0.759 & 0.725 & 0.742 \\
DNN & 0.736 & 0.730 & 0.733 \\
KAN (Width: [50, 25, 1], K: 15, Grid: 5) & 0.714 & 0.624 & 0.666 \\
\hline
\end{tabular}
}
\end{resizebox}
\caption{Performance Metrics of Various Classifiers for Dataset 5. KAN demonstrated moderate performance but was outperformed by classifiers like Random Forest, Gradient Boosting, and LightGBM, which achieved higher F1 scores.}
\label{table:classifier_metrics_dataset5}
\end{table}

\begin{figure}[H]
    \centering
    \includegraphics[width=\linewidth]{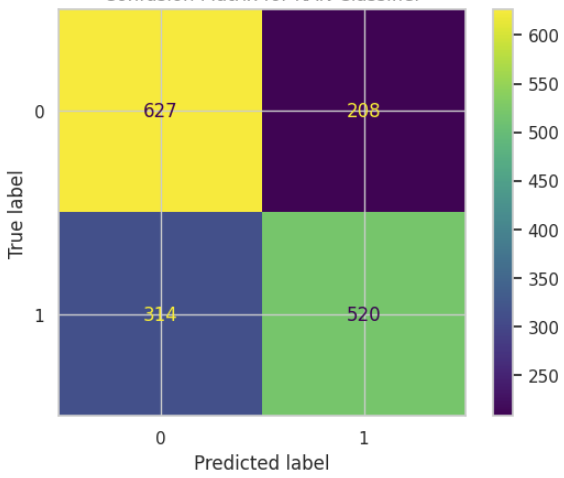}
    \caption{KAN Confusion Matrix for Dataset 5}
    \label{fig:kan_confusion_matrix_5}
\end{figure}

\bibliographystyle{unsrt}
\bibliography{reference}

\end{document}